# Deep Learning for Recognizing Mobile Targets in Satellite Imagery

Mark Pritt
Lockheed Martin Space
Rockville, Maryland
mark.pritt@lmco.com

***Abstract*—There is an increasing demand for software that automatically detects and classifies mobile targets such as airplanes, cars, and ships in satellite imagery. Applications of such automated target recognition (ATR) software include economic forecasting, traffic planning, maritime law enforcement, and disaster response. This paper describes the extension of a convolutional neural network (CNN) for classification to a sliding window algorithm for detection. It is evaluated on mobile targets of the xView dataset, on which it achieves detection and classification accuracies higher than 95%.**

***Keywords—ATR; target recognition; artificial intelligence; AI; deep learning; CNN; neural networks; machine learning; image understanding; recognition; classification; satellite imagery***

## I. Introduction

Deep learning is a class of machine learning algorithms that have enjoyed astonishing success in object detection and classification [1] and automated target recognition (ATR). They have achieved this success by combining the computational power of deep convolutional neural networks (CNNs) with the processing power of graphical processing units (GPUs). Furthermore, unlike conventional ATR methods, CNNs do not require the algorithm designer to engineer feature detectors. The network itself learns which features to detect, and how to detect them, as it trains.

Much work has been aimed at the problem of classification, which is the process of assigning a label or class to an image of a single object as shown in Fig. 1(a). The Intelligence Advanced Research Projects Agency (IARPA) recently sponsored a TopCoder competition called fMoW or Functional Map of the World [2,3] in which competitors developed classification algorithms for a global dataset of satellite images containing fixed facilities. The facilities were of 62 types or classes and included airports, shipyards, schools, stadiums, and towers as shown in Fig. 2(a). The high-resolution images came from the WorldView-2 satellite and had a ground sample distance (GSD) of 0.5m. The winning algorithms achieved classification accuracies approaching 80% [4].

Detection is a more complicated task than classification. It is the process of recognizing multiple objects in an image and classifying them as shown in Fig. 1(b). (This paper uses the term "detection" in the broad sense of detecting and classifying objects. It is essentially a synonym for recognition.) The National Geospatial-Intelligence Agency (NGA) and Defense Innovation Unit (DIU) recently sponsored a coding challenge called xView [5,6] in which competitors developed detection algorithms for a global dataset of satellite images containing mobile targets. The images were from the WorldView-3 satellite and had a GSD of 0.3m. The targets consisted of 60 classes of aircraft, motor vehicles, maritime vessels, construction equipment, and railway vehicles, as well as some fixed facilities, as shown in Fig. 2(b). See Table I for a complete list of the classes.

The winners of the xView competition developed algorithms whose best metrics were much lower than those of the fMoW algorithms. The mean average precision (mAP), a common metric of detection success, did not exceed 30% [7]. Nor did the $F_1$ score, a measure of accuracy and precision. These results testify to the difficulty of detecting mobile targets in satellite imagery.

## II. Problem

There is an increasing demand for the recognition of mobile targets in satellite imagery. Car counting in satellite images is used by investors to estimate the retail activity of large businesses. It is also used to monitor traffic for purposes of road planning and traffic flow management. Aircraft detection is used to estimate air traffic volume at airports. Maritime surveillance is used for monitoring ship traffic, detecting illegal fishing activity, and protecting borders from drug runners and human traffickers. Even disaster response is an important application of mobile target detection as indicated by the xView competition [6].

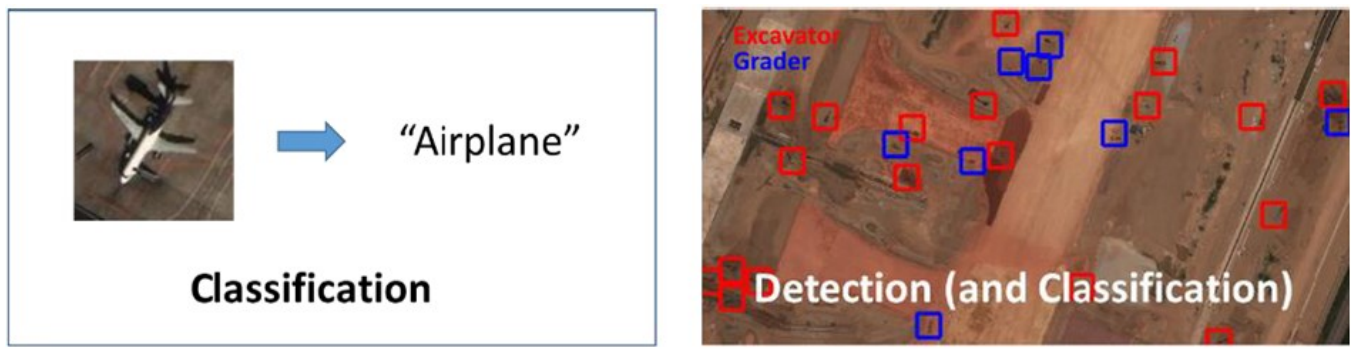

(a) (b)

Fig. 1. (a) The process of classification assigns a label to an image of an object. (b) The process of detection recognizes multiple objects in an image.

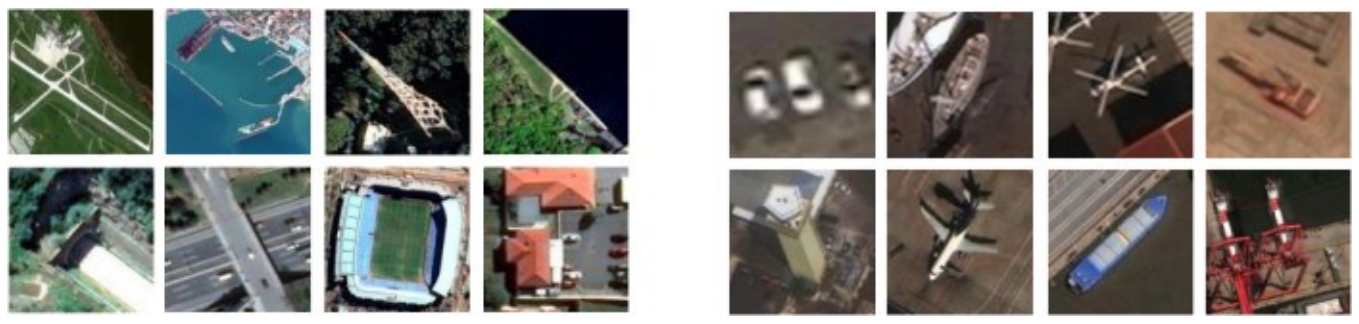

(a) (b)

Fig. 2. (a) The fMoW (Functional Map of the World) dataset consists of images of fixed facilities such as bridges, towers, stadiums and road intersections. (b) The xView dataset consists mainly of mobile targets.

TABLE I. CLASSES OF THE XVIEW DATASET. EACH CLASS HAS AN IDENTIFICATION NUMBER AND A LABEL.

| | |
|---|---|
| **Aircraft** | **Railway** |
| 11: Fixed-wing Aircraft | 33: Railway Vehicle |
| 12: Small Aircraft | 34: Passenger Car |
| 13: Passenger/Cargo Plane | 35: Cargo/Container Car |
| 15: Helicopter | 36: Flat Car |
| | 37: Tank Car |
| **Motor Vehicles** | 38: Locomotive |
| 17: Passenger Vehicle | |
| 18: Small Car | **Maritime** |
| 19: Bus | 40: Maritime Vessel |
| 20: Pickup Truck | 41: Motorboat |
| 21: Utility Truck | 42: Sailboat |
| 23: Truck | 44: Tugboat |
| 24: Cargo Truck | 45: Barge |
| 25: Truck Tractor w/ Box Trailer | 47: Fishing Vessel |
| 26: Truck Tractor | 49: Ferry |
| 27: Trailer | 50: Yacht |
| 28: Truck Tractor w/ Flatbed Trailer | 51: Container Ship |
| 29: Truck Tractor w/ Liquid Tank | 52: Oil Tanker |
| 32: Crane Truck | |
| | **Fixed Facilities** |
| **Construction Vehicles** | 71: Hut/Tent |
| 53: Engineering Vehicle | 72: Shed |
| 54: Tower Crane | 73: Building |
| 55: Container Crane | 74: Aircraft Hangar |
| 56: Reach Stacker | 76: Damaged Building |
| 57: Straddle Carrier | 77: Facility |
| 59: Mobile Crane | 79: Construction Site |
| 60: Dump Truck | 83: Vehicle Lot |
| 61: Haul Truck | 84: Helipad |
| 62: Scraper/Tractor | 86: Storage Tank |
| 63: Front loader/Bulldozer | 89: Shipping Container Lot |
| 64: Excavator | 91: Shipping Container |
| 65: Cement Mixer | 93: Pylon |
| 66: Ground Grader | 94: Tower |

Because the mobile targets to be detected are so numerous, automated detection and classification algorithms are required. Yet traditional ATR algorithms are inaccurate and unreliable. What is needed is an advanced deep learning system that can recognize and label mobile targets automatically. Such a system must do so with metrics much higher than the poor results achieved by the xView competition.

This paper presents a start in that direction. It describes an algorithm for the recognition of mobile targets in satellite imagery with an accuracy of 95% or higher. It focuses on the Passenger/Cargo Plane class of aircraft as an important class of mobile target. It also focuses on the Excavator and Ground Grader classes as examples of particularly small mobile targets that are difficult to detect and classify in satellite imagery.

## III. DATASET

The xView dataset [6] was chosen for the experiments because it contains a wide variety of mobile targets. The dataset consists of approximately 1000 color satellite images from around the world with about 600,000 objects identified and labeled. The images, shown in Fig. 3, are orthorectified and from the WorldView-3 satellite. They are approximately 3000x3000 pixels in size and have a GSD of 0.3 meters. The objects are classified into the 60 classes listed in Table I. Forty-six of the classes are mobile targets and grouped into motor vehicles, construction equipment, railway vehicles, aircraft, and maritime vessels. The objects are identified by means of bounding boxes and labels in a GeoJSON file, whose structure is shown in Table II.

The xView dataset is unparalleled as the largest global dataset of mobile targets in satellite imagery. It has been criticized, however, for its class ambiguities and inconsistent labeling [8]. For example, there are maritime vessels that are not labeled as Maritime Vessel. There is a Vehicle Lot class, but vehicles in parking lots are not always labeled with this class. It is not clear why some classes, such as Passenger Car or Tank Car, are not also classified as Railway Vehicle. The dataset also suffers from a very large class imbalance. There are 316,138 examples of the Building class and 210,938 examples of the Small Car class but only 55 examples of Helicopter and 17 examples of Railway Vehicle.

TABLE II. XVIEW GEOJSON FILE STRUCTURE

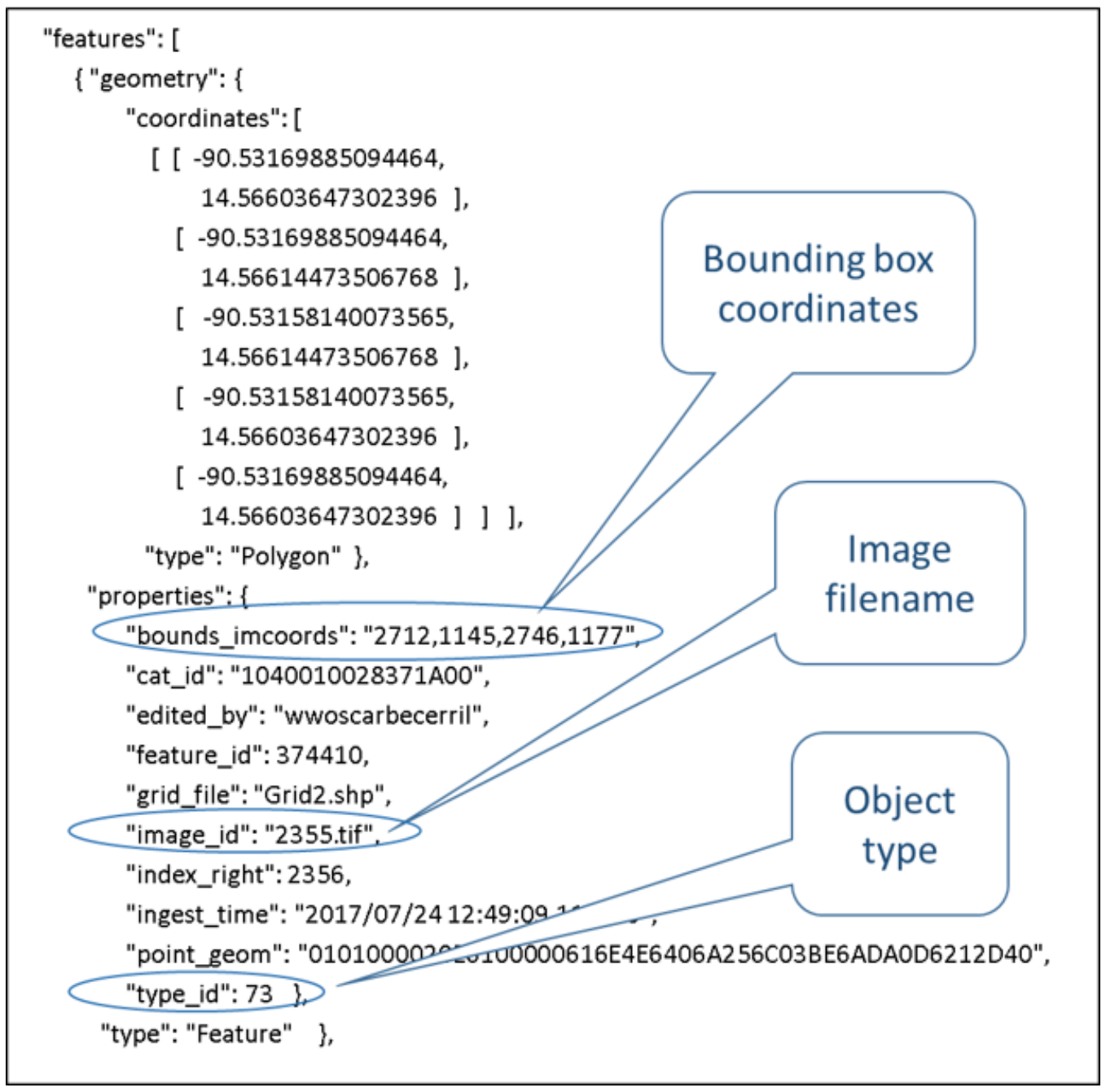

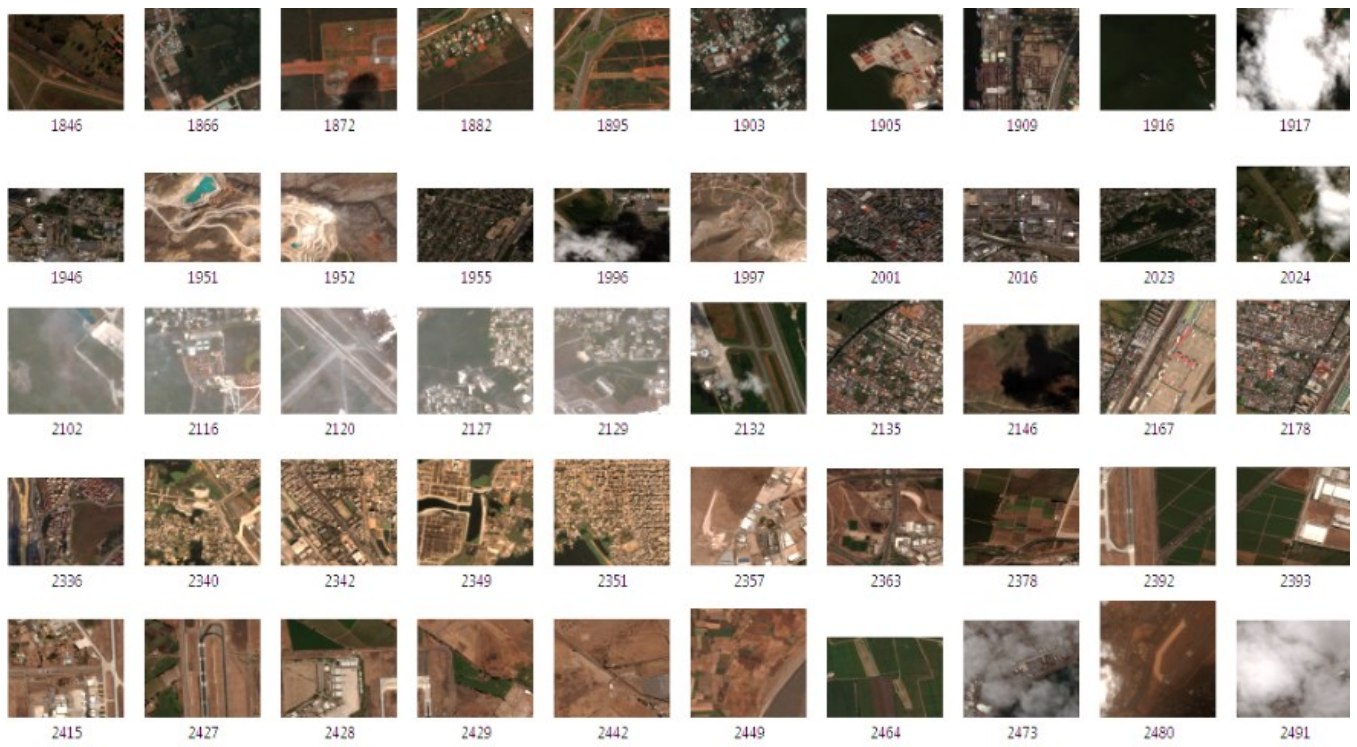

Fig. 3. Some of the 1000 satellite images from the xView dataset.

## IV. METHODS

The goal of the work described in this paper was to develop a deep learning algorithm for the detection and classification of mobile targets in satellite imagery. A classifier on the 60 xView classes was first trained to assess the quality of the dataset. A binary aircraft classifier was then trained and integrated in a sliding window detector, which detected and classified aircraft in satellite images. An excavator and ground grader classifier was then trained and inserted in the sliding window detector. This demonstrated the ability of the algorithm to detect and classify small mobile targets.

### A. Metrics

The most common metric for evaluating detection and classification algorithms is *accuracy*. It is the probability of detecting and classifying an object into its correct class. Also called $P_d$, *probability of detection*, or *recall*, it is the proportion of a given class that is correctly detected and classified. *Precision* is a complementary measure that is defined to be the proportion of classifications of a given class that are correct. If *TP* is the number of true positives, *FP* the false positives, and *FN* the false negatives, then the precision *P* and recall *R* are defined by the equations

$$P = TP/(TP + FP), \qquad R = TP/(TP + FN).$$

A *confusion matrix* expresses the precision and recall of all the classes. It is a table that records the class accuracies along the diagonal and the likelihood of class confusion on the off-diagonal elements. Another metric is the $F_1$ *score*, which is the geometric mean of the precision and recall: $2PR/(P + R)$. Finally, mAP or *mean average precision* is a measure that is commonly used for detection. It measures the precision and also the effectiveness of localization (i.e., accurate placement of the detection bounding boxes.)

This paper uses precision, recall, $F_1$ score, and confusion matrices to measure the effectiveness of the algorithms.

### B. xView Classification

To assess the xView dataset, a multi-class CNN classifier was trained and evaluated. This was deemed necessary to determine if classification accuracies significantly higher than the low metrics (30%) of the xView competition could be achieved. It was considered important because of the dataset's weaknesses of inconsistent labeling and class imbalance.

As with the Lockheed Martin fMoW classifier [9], a new class called "False Detection" was defined with the purpose of teaching the CNN what the 60 xView classes did *not* look like. This new class was defined by sampling the xView images (Fig. 3) at random locations and extracting bounding boxes at the locations, which were then added to the GeoJSON file. As in [9], the training data was split 80%-20% into a training dataset, used for training the CNN, and a validation dataset, which was used only for calculating the metrics of the CNN: precision, recall, and $F_1$ score. To mitigate the class imbalance of xView, the number of examples per class was capped at 10,000, and each class was weighted according to its training set size. The training data was augmented by means of flips and 90-degree rotations as in [9], in addition to random image shifting and re-sizing.

A DenseNet-161 CNN [10] was implemented in Python with the TensorFlow [11] and Keras [12] deep learning libraries. The CNN weights were initialized with ImageNet weights as in [9] and trained for one epoch with a learning rate of 0.0005, dropout 0.6, and GPU batch size 16.

When evaluated on the validation dataset, the CNN classifier exhibited a total accuracy of 73%. Fig. 4 shows the class accuracies. They ranged from a high of 100% for the classes of Straddle Carrier and Helipad to a low of 28% for Utility Truck and 21% for Engineering Vehicle. The accuracies of 9 classes exceeded 90%. Fig. 5 shows the confusion matrix. The strong diagonal indicates a strong classifier, with the off-diagonal elements indicating classes that were confused with others. The most common confusion was between Tower and Storage Tank (33%). Passenger Vehicle was confused with Small Car 31% of the time, but this was probably due to labeling ambiguities in the training data. Ground Grader was confused with Front Loader/Bulldozer 25% of the time, most likely because of their similar appearances.

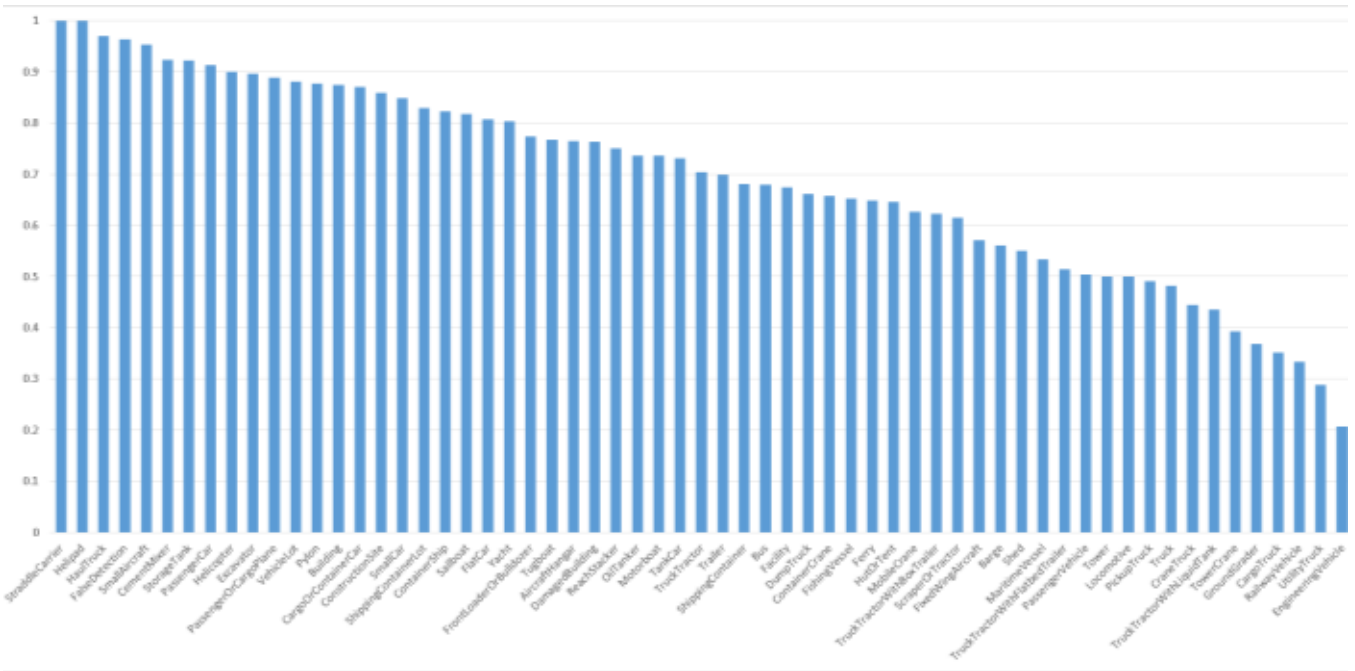

Fig. 4. Class accuracies of the 61-class xView classifier. The classes consisted of the 60 xView classes and a False Detection class.

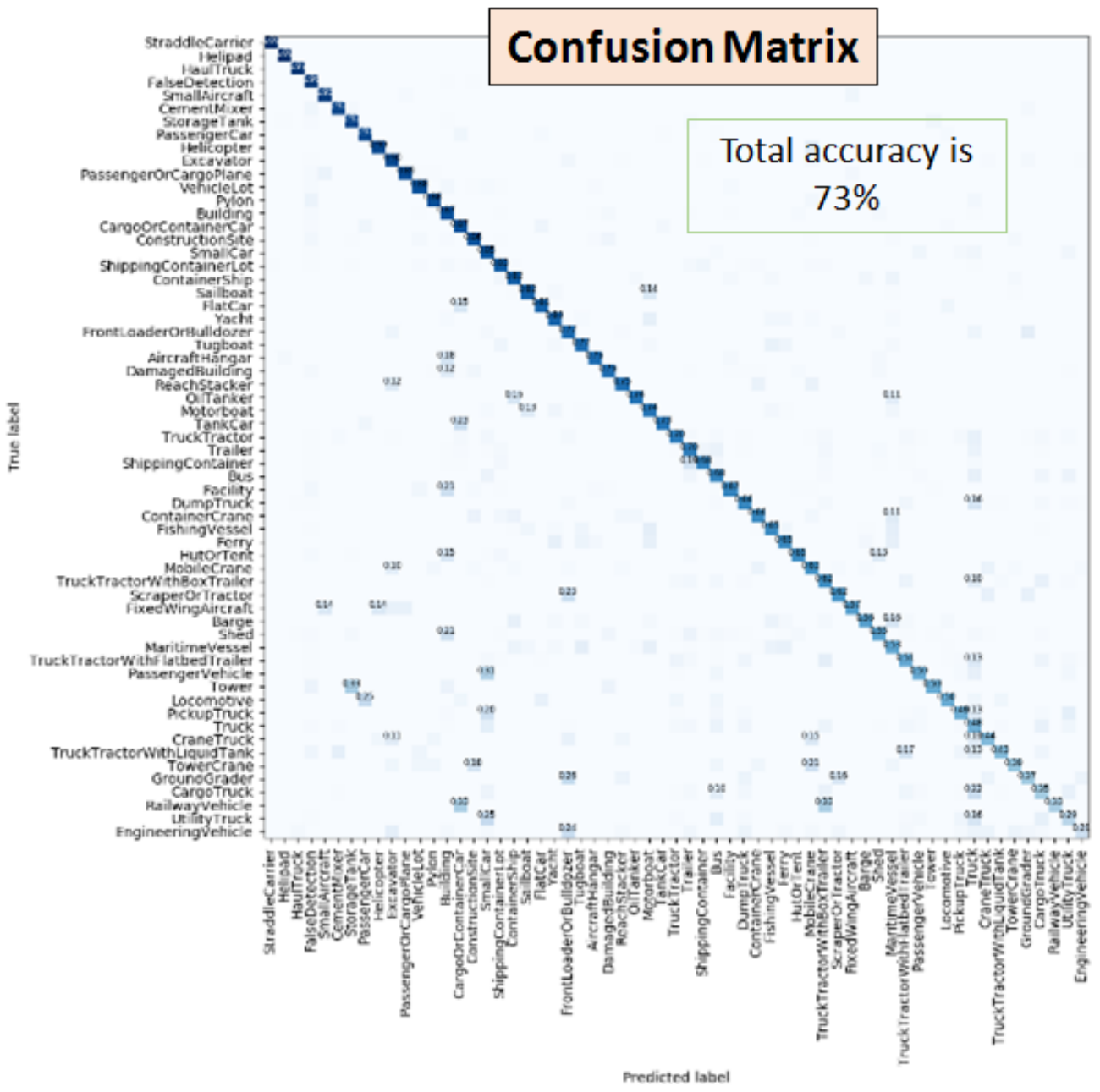

Fig. 5. Confusion matrix of the 61-class xView classifier (including the False Detection class).

The precision and recall of the Passenger/Cargo Plane class were 98% and 89%, respectively. Those of the Excavator class were 80% and 90%, while those of the Ground Grader class were only 24% and 37%.

### C. *Aircraft Classification*

Next a binary classifier was trained to recognize airplanes and distinguish them from background clutter. A ResNet-50 CNN [13] was trained on the two classes of Passenger/Cargo Plane (Fig. 6) and False Detection. (The DenseNet-161 CNN [10] yielded similar results.) There were 719 sample images of Passenger/Cargo Plane in the xView dataset, and the usual 80%-20% dataset split was performed along with augmentation of the training images by means of flips, rotations, shifts, and scale changes. The CNN weights were initialized with ImageNet weights as in [9] and trained for two epochs with a learning rate of 0.0005, dropout 0.6, and GPU batch size 16.

When evaluated on the validation dataset, the binary classifier exhibited an accuracy of 97% and a precision of 94% as shown in Table III. The accuracy was 8% higher than that of the aircraft in the xView multi-class classifier. The failures are shown in Fig. 7. Several were caused by the false classification of small aircraft. Two failures were actually not failures at all, because they were caused by human labeling mistakes in the validation dataset. Moreover, a passenger plane in a hangar building with only its tail exposed was correctly classified even though it had been left unlabeled in the dataset.

### D. *Sliding Window Detection*

For the detection algorithm, a sliding window detector was implemented. This algorithm consisted of the following steps:

1. Divide the input image into overlapping windows of a predetermined size as shown in Fig. 8.
2. Submit each window to the classifier and save the windows whose classification (prediction) probabilities exceed a predetermined threshold.
3. Apply a non-maximal suppression algorithm to remove overlapping windows, ensuring that each object is detected only once. The windows that remain are output as the detections.

The advantages of this algorithm are its simplicity and the natural way it extends the classifier. It is a simple matter to "plug and play" a different classifier in step 2, such as an ensemble [9]. The main disadvantage of the algorithm is its relatively slow speed compared to one-pass algorithms such as Faster R-CNN [14]. It is possible, however, to speed it up by means of a selective search [15] or pre-screening algorithm. Another disadvantage is that the algorithm tends to have poor localization for aircraft. In other words, the detection windows are not always centered on the airplanes they contain.

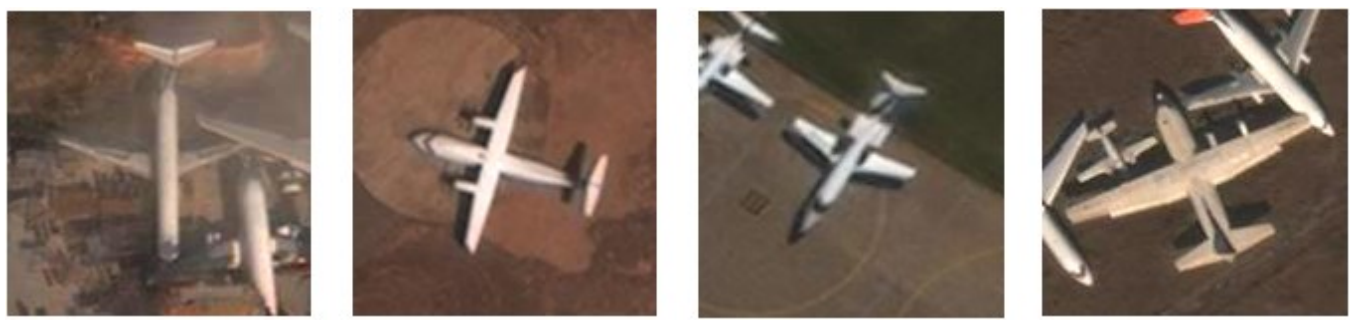

Fig. 6. Examples of the "Passenger/Cargo Plane" class in the xView dataset.

TABLE III. AIRCRAFT CLASSIFICATION RESULTS

| Metric | Value |
|---|---|
| Recall (Accuracy) | 97% |
| Precision | 94% |
| $F_1$ Score | 96% |
| True Positives | 111 |
| False Positives | 3 |
| False Negatives | 7 |
| True Negatives | 2818 |

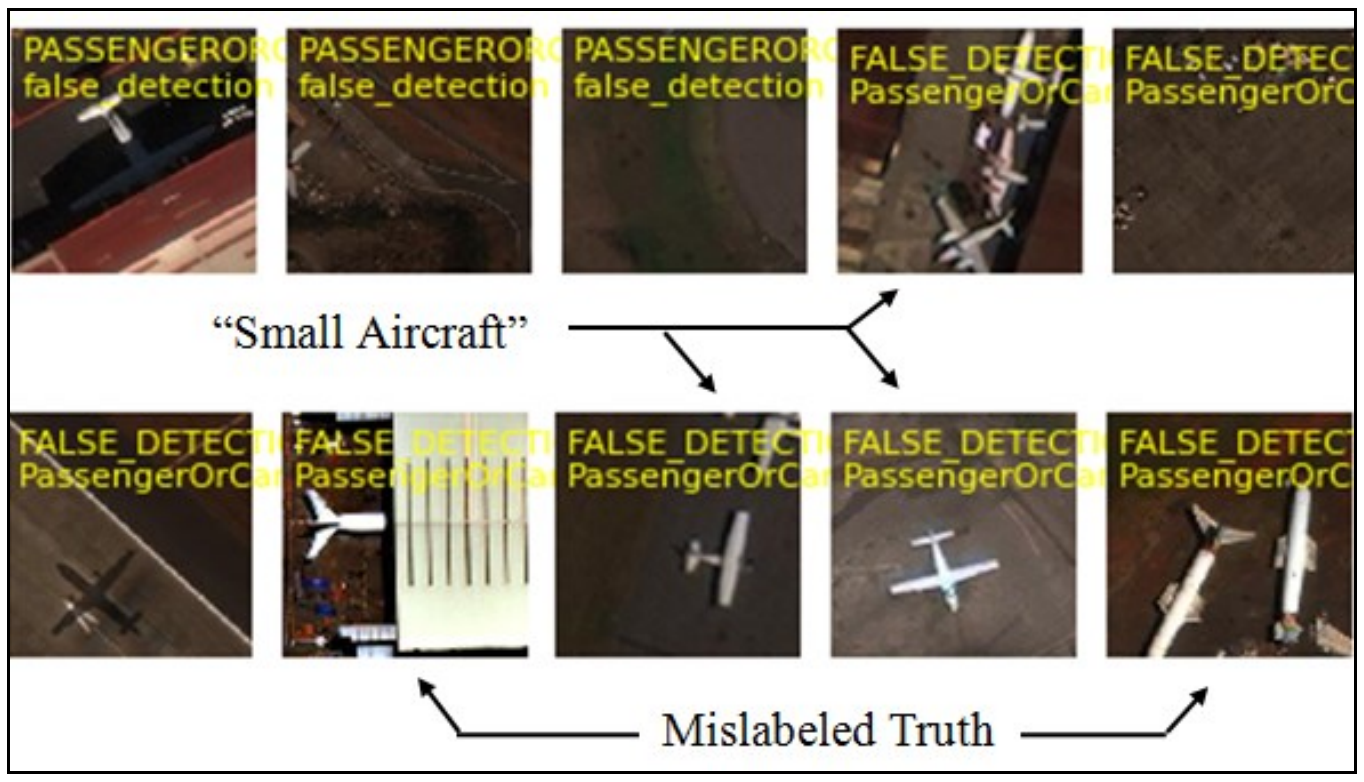

Fig. 7. Failures of the aircraft classifer.

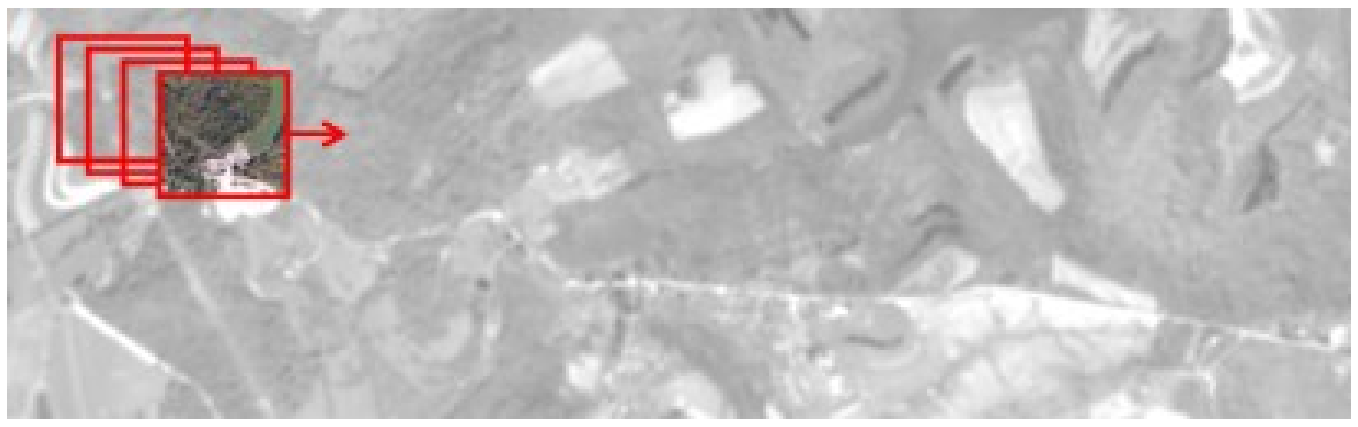

Fig. 8. Sliding window algorithm.

Fig. 9 shows the results of the sliding window algorithm for aircraft detection. The red boxes are the detection windows that resulted from step 3. Note the poor localization of some of the windows, as well as the false positive detection. The latter resulted from the presence of the wing tips of two airplanes in the window, which provided enough features to trigger the CNN with a high prediction probability. It is possible that a better non-maximal suppression algorithm would remove this false detection and exhibit better localization.

The size of the windows in step 1 was determined by the sizes of the bounding boxes in the training dataset. In general, it could be set to the mean size or a set of sizes, in which case step 1 would consist of two or more passes. In the case of aircraft, which can be small (e.g., a corporate jet) or large (e.g., a jumbo jet), it was set to two different sizes: the mean size and the $75^{th}$ percentile of the bounding box sizes.

The probability threshold in step 2 was determined by the prediction probabilities recorded in the validation step of the CNN classifier training. It was set to $\mu$ - $3\sigma$, where $\mu$ was the mean of these probabilities and $\sigma$ was the standard deviation.

## E. Excavator Classification and Detection

A binary classifier was next trained to recognize excavators. The xView dataset has 846 examples of this class. Fig. 10 shows an excavator and four examples of excavators in satellite imagery. A ResNet-50 CNN was trained on the two classes of Excavator and False Detection. The usual 80%-20% dataset split was performed along with augmentation of the training data by means of flips, rotations, shifts, and scale changes. The CNN weights were initialized as usual with ImageNet weights and trained for two epochs with a learning rate of 0.0005, dropout 0.6, and GPU batch size 16.

When evaluated on the validation dataset, the classifier exhibited an accuracy of 94% and a precision of 91% as shown in Table IV. When implemented in the sliding window detector, it was noted that excavators were often confused with ground graders, which are shown in Fig. 11. A three-class CNN classifier was then trained on Excavator, Ground Grader, and False Detection. Even though xView has only 83 examples of the Ground Grader class, the resulting detector achieved significantly better results as shown in the following section.

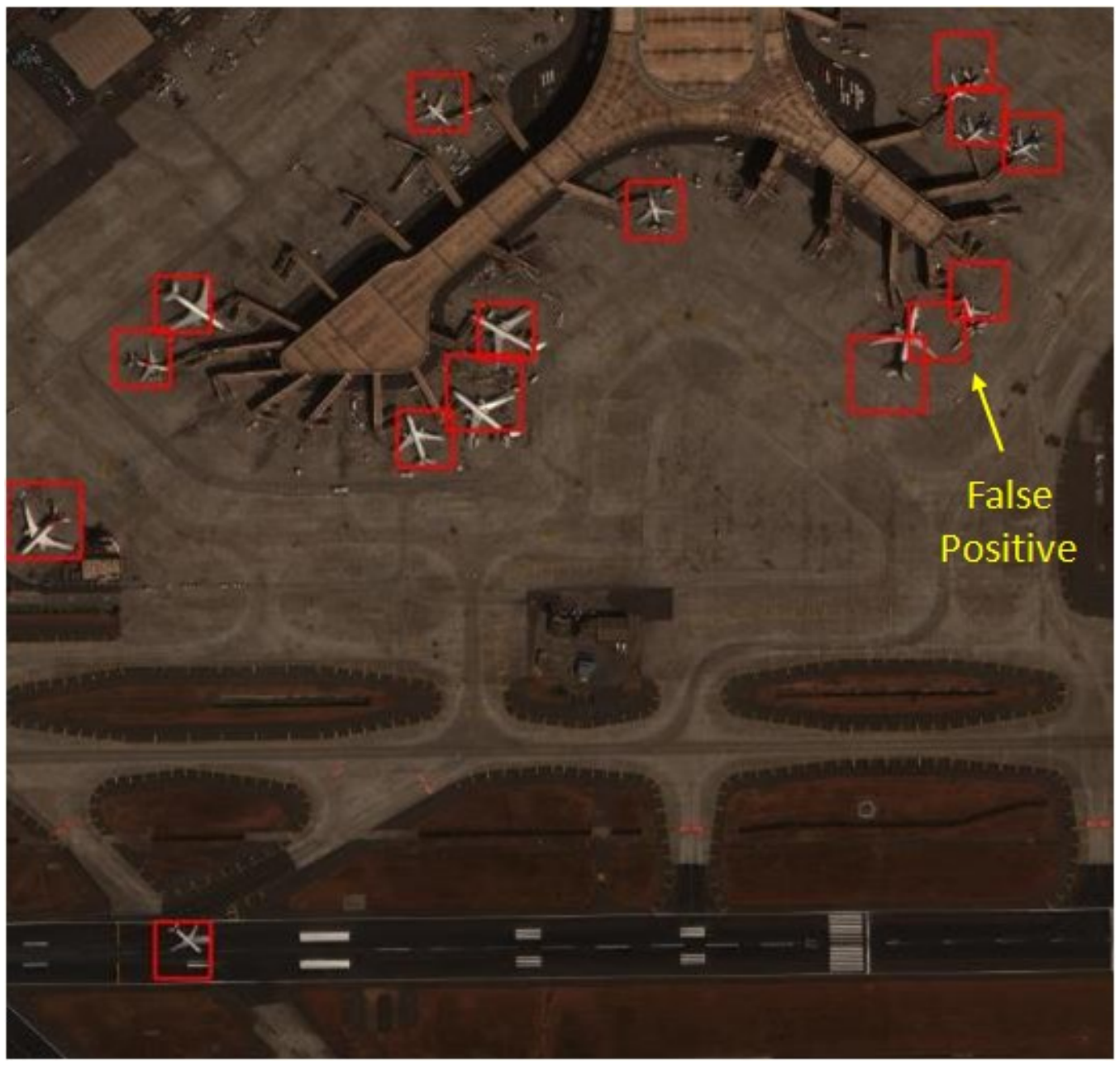

Fig. 9. The red boxes are the detection results. All the aircraft were detected, but there was a false positive detection straddling two airplanes.

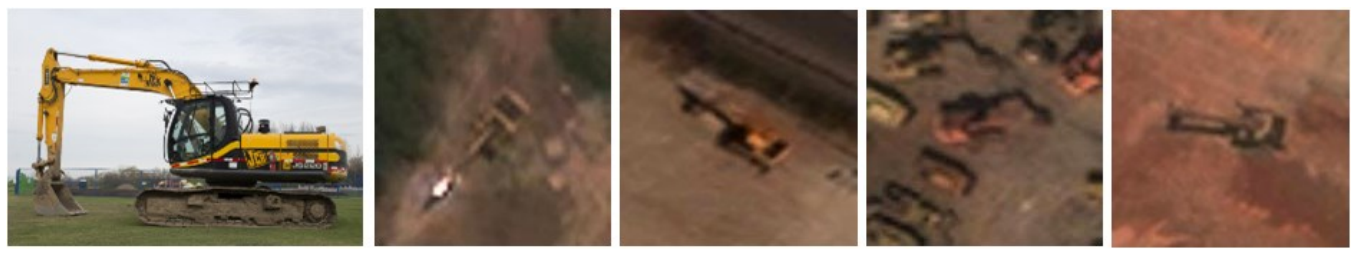

Fig. 10. The photo on the left is an excavator, and the other photos are satellite images of excavators.

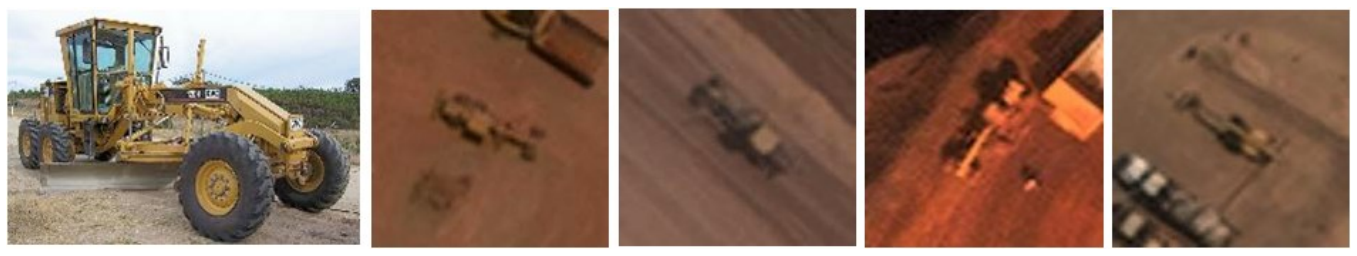

Fig. 11. The photo on the left is a ground grader, and the other photos are satellite images of ground graders.

# V. Results and Discussion

Table V shows the results of the sliding window detector on the Passenger/Cargo Plane class. The algorithm detected aircraft with an accuracy of 96% and a precision of 92%. The detection of aircraft under hazy visibility is shown in Fig. 12, and Fig. 13 shows the ability of the algorithm to reject confusing look-alikes: in this case, jetways that resemble airplane wings.

Table VI shows the detection results on the Excavator class. The accuracy was only 89%. It was noted that many of the false detections were ground graders, which appear similar to excavators. Fig. 14 shows the detections on an image of a large airport construction site. The three false positives in the detail image were all ground graders.

When the CNN classifier was retrained on the three classes (Excavator, Ground Grader, and False Detection) and plugged into the sliding window algorithm in step 2, the detection results improved substantially. Table VII shows the accuracy increasing from 89% to 95% and the precision increasing from 81% to 91%. The results on the airport construction site are shown in Fig. 15. The false positives and false negatives disappeared in the detail image.

Interestingly, in both cases of aircraft and excavator, the accuracy and precision of the detector generally matched those of the classifier.

# VI. Conclusion

In this paper a sliding window algorithm was presented for the detection and classification of mobile targets in satellite imagery. The xView dataset was used to test the algorithm, which achieved accuracies of 95% and 96% on the aircraft and excavator classes. These metrics are higher than those achieved by the xView competition. Although slow and sometimes poor at aircraft localization, the algorithm enjoys the advantages of simplicity and the ability to substitute any classifier in “plug and play” fashion.

TABLE IV. Excavator Classification Results

| Metric | | Value | |
|---|---|---|---|
| Recall (Accuracy) | | 94% | |
| Precision | | 91% | |
| $F_1$ Score | | 92% | |
| True Positives | 137 | False Positives | 9 |
| True Negatives | 3314 | False Negatives | 14 |

TABLE V. Aircraft Detection Results

| Metric | Value |
|---|---|
| Recall (Accuracy) | 96% |
| Precision | 92% |
| $F_1$ Score | 94% |
| True Positives | 76 |
| False Positives | 7 |
| False Negatives | 3 |

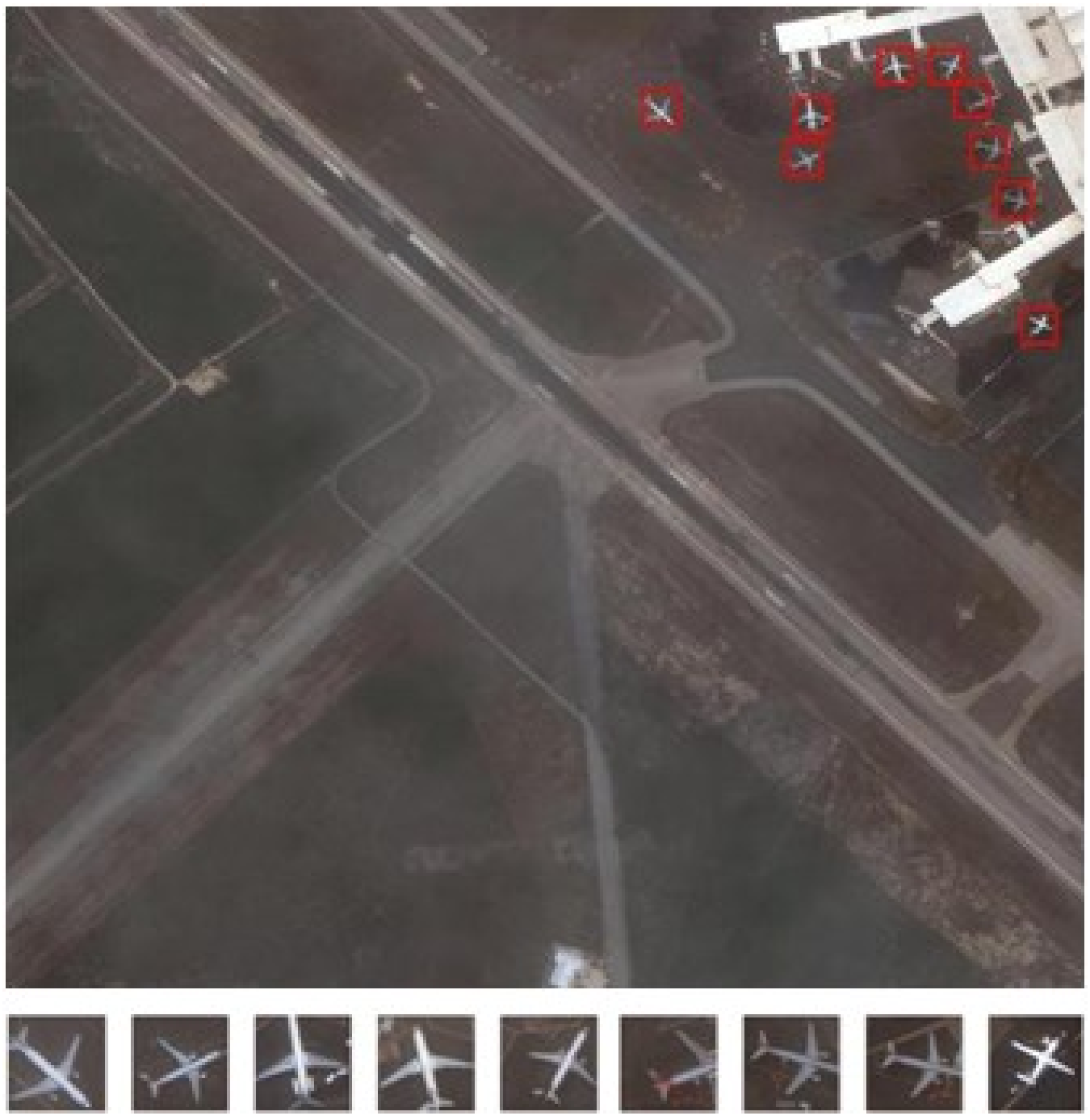

Fig. 12. Detection results on a hazy image. All nine airplanes were detected.

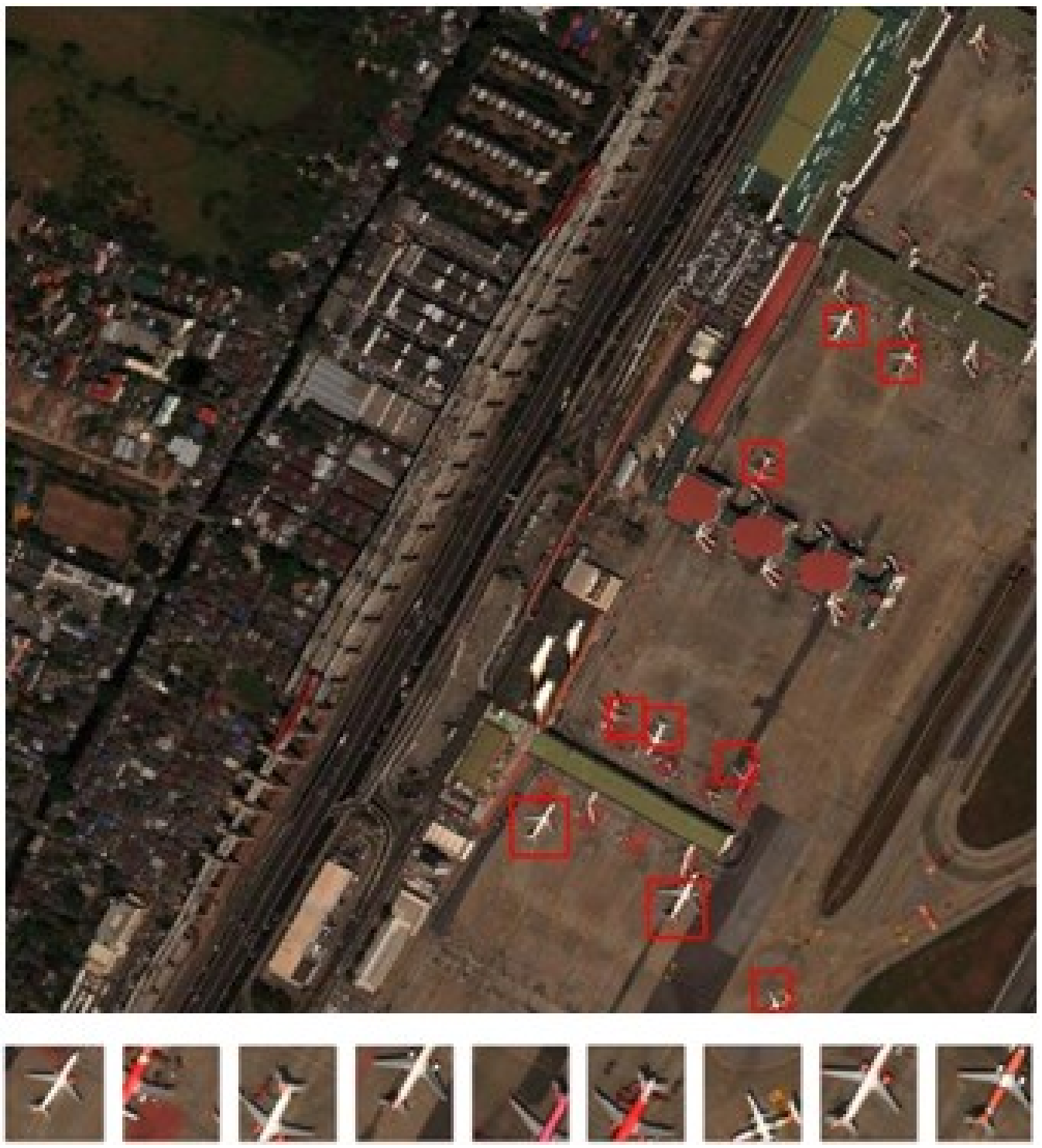

Fig. 13. All nine airplanes were detected with no false positives. These results show the ability of the algorithm to reject confusing look-alikes (jetways).

TABLE VI. EXCAVATOR DETECTION RESULTS

| Metric | Value |
|---|---|
| Recall (Accuracy) | 89% |
| Precision | 81% |
| $F_1$ Score | 85% |
| True Positives | 50 |
| False Positives | 12 |
| False Negatives | 6 |

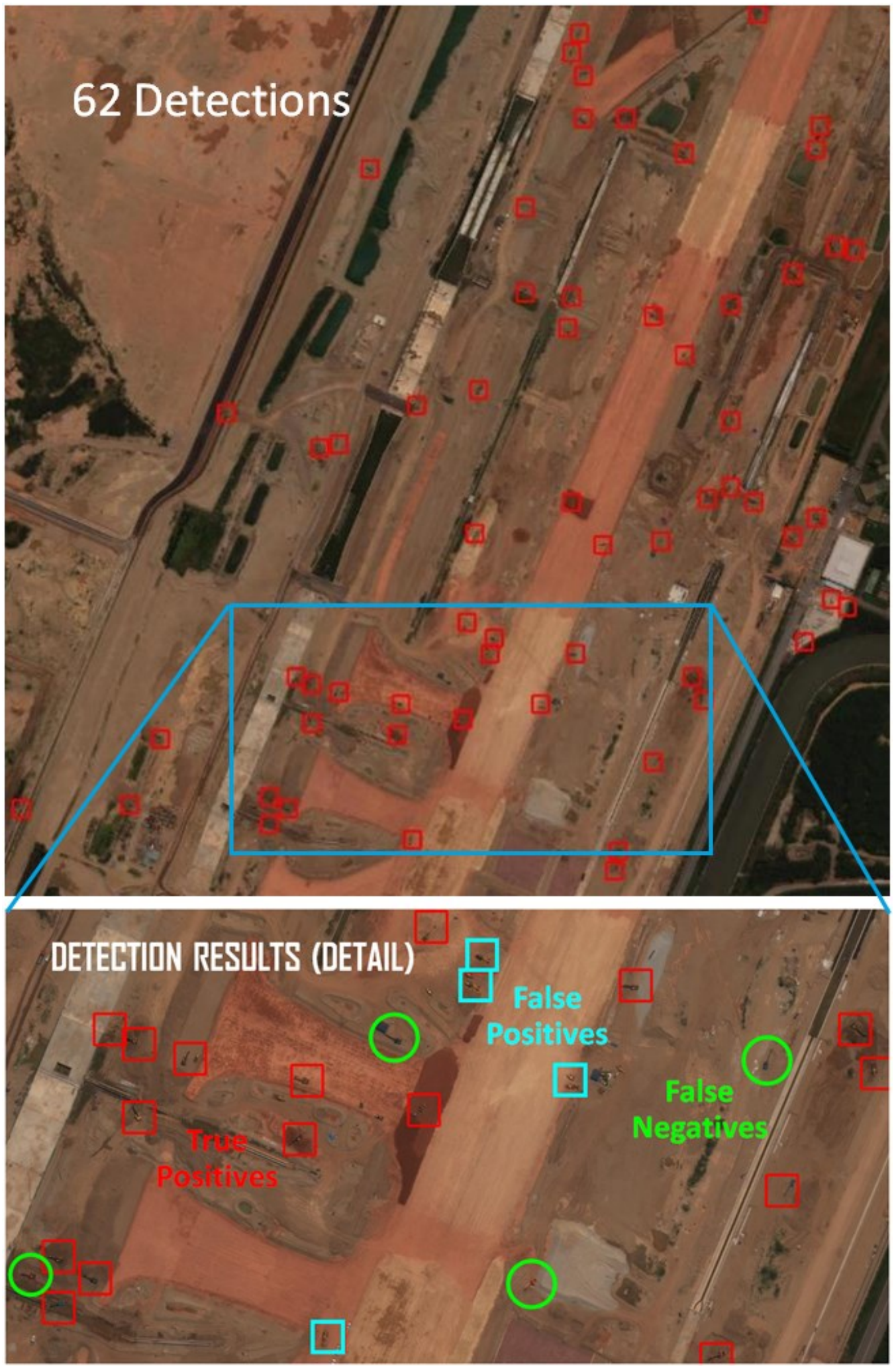

Fig. 14. The top image shows the detection results on a satellite image of an airport construction site. The red boxes indicate the detections of excavators. The bottom image shows a portion of the image in magnified detail. Three of the detections were ground graders instead of excavators (false positives) and four of the excavators were not detected (false negatives).

There are a number of ways the results could be improved. First, the non-maximal suppression of the detection windows in step 3 could be modified to provide more accurate object localization and better detection of objects that are crowded together. Second, a method could be devised to determine automatically the optimal set of target classes on which the classification CNN should be trained. This optimal set could be determined, for example, by the highest values of the off-diagonal elements in the classifier's confusion matrix. Third, the classification CNN could be replaced with an ensemble of CNNs [9] for higher classification accuracy, which would in turn produce better detection metrics. Finally, the sliding window algorithm could be compared with one-pass detection algorithms such as Faster R-CNN [14].

## ACKNOWLEDGEMENTS

The author would like to acknowledge Gary Chern and Ryan Soldin for helpful discussions about deep learning for object detection.

## REFERENCES AND NOTES

[1] Y. LeCun, Y. Bengio, and G. Hinton, "Deep learning," Nature, vol. 521, pp. 436-444, 28 May 2015.

[2] "Functional Map of the World Challenge", IARPA, https://www.iarpa.gov/challenges/fmow.html.

[3] G. Christie. N. Fendley, J. Wilson, and R. Mukherjee, "Functional map of the world," arXiv 1711.07846, 21 Nov 2017.

[4] "Marathon Match: Functional Map of the World", TopCoder, https://community.topcoder.com/longcontest/stats/?module=ViewOverview&rd=16996.

[5] "DIUx xView 2018 Detection Challenge", Defense Innovation Unit, http://xviewdataset.org/.

[6] D. Lam et al., "xView: objects in context in overhead imagery," arXiv 1802.07856, 22 Feb 2018.

[7] Mean average precision (mAP) is a standard metric for object detection. It is very roughly comparable to accuracy. The highest mAP achieved in the xView challenge was 29% and the highest $F_1$ score, which is the geometric mean of accuracy and precision, was 30%. These results were discussed by B. McCord, "xView challenge satellite objects in context", IEEE Workshop Applied Imagery Pattern Recognition (AIPR), Oct 2018. The xView results were officially announced by B. McCord, "Advancing computer vision frontiers for disaster response: the xView detection challenge," GPU Technology Conference, Washington, DC, 22-24 Oct 2018.

[8] R. Fong, "Xview dataset and baseline results," Medium, 25 May 2018, https://medium.com/picterra/the-xview-dataset-and-baseline-results-5ab4a1d0f47f.

[9] M. Pritt and G. Chern, "Satellite image classification with deep learning," IEEE Workshop Applied Imagery Pattern Recognition (AIPR), Oct 2017.

[10] G. Huang, "Dense connected convolutional neural networks," IEEE Computer Society Conference on Computer Vision and Pattern Recognition (CVPR), 2017.

[11] "TensorFlow: An open-source software library for machine intelligence," TensorFlow, https://www.tensorflow.org/.

[12] F. Chollet at al., "Keras", GitHub, 2017, https://github.com/fchollet/keras.

[13] K. He et al., "Deep residual learning for image recognition," arXiv 1512.03385, Dec 2015.

[14] S. Ren, K. He, R. Girshick and J. Sun, "Faster R-CNN: towards real-time object detection with region proposal networks," arXiv 1506.01497v3 [cs.CV], 6 Jan 2016.

[15] J. Uijlings, K. Sande, T. Gevers and A. Smeulders, "Selective search for object recognition," Int. J. Computer Vision, vol. 104, pp. 154-171, 2013.

TABLE VII. IMPROVED EXCAVATOR DETECTION RESULTS

| Metric | Value |
|---|---|
| Recall (Accuracy) | 95% |
| Precision | 91% |
| $F_1$ Score | 93% |
| True Positives | 53 |
| False Positives | 5 |
| False Negatives | 3 |

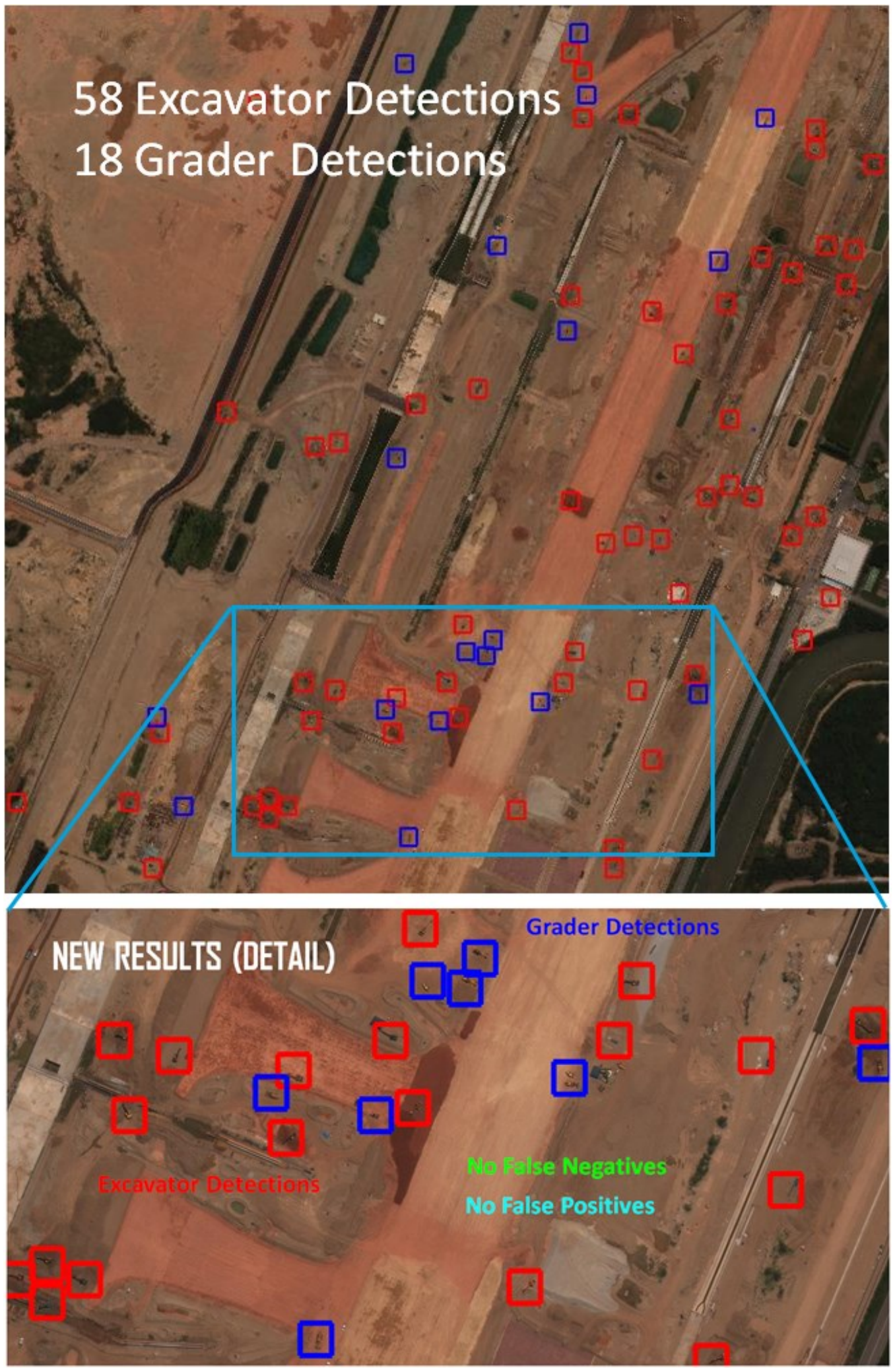

Fig. 15. Improved detection results on the airport construction site.